\documentclass[11pt]{article}

\usepackage[final]{acl}

\usepackage{times}
\usepackage{latexsym}

\usepackage[T1]{fontenc}
\usepackage[utf8]{inputenc}

\usepackage{microtype}

\usepackage{inconsolata}

\usepackage{graphicx}

\usepackage{amsmath}
\usepackage{algpseudocode}
\usepackage{makecell}
\newtheorem{principle}{Principle}
\usepackage{adjustbox}
\usepackage{booktabs}
\usepackage{multirow}

\DeclareMathOperator*{\concat}{\mathbin{\|}}

\title{Interpretability from the Ground Up: Stakeholder-Centric Design of Automated Scoring in Educational Assessments}
\author{
  Yunsung Kim$^1$, Michael Hardy$^1$, Joseph Tey$^2$, Candace Thille$^1$, Chris Piech$^2$ \\
  $^1$Graduate School of Education, Stanford University \\
  $^2$Department of Computer Science, Stanford University \\
  \texttt{\{yunsung,hardym,joetey,cthille,cpiech\}@stanford.edu}
}

\begin{document}
\maketitle
\begin{abstract}
AI-driven automated scoring systems offer scalable and efficient means of evaluating complex student-generated responses. Yet, despite increasing demand for transparency and interpretability, the field has yet to develop a widely accepted solution for interpretable automated scoring to be used in large-scale real-world assessments. This work takes a principled approach to address this challenge. We analyze the needs and potential benefits of interpretable automated scoring for various assessment stakeholder groups and develop four principles of interpretability -- \textbf{F}aithfulness, \textbf{G}roundedness, \textbf{T}raceability, and \textbf{I}nterchangeability (\textbf{FGTI}) -- targeted at those needs. To illustrate the feasibility of implementing these principles, we develop the \textsc{AnalyticScore} framework\footnote{Implementation of \textsc{AnalyticScore} is available at: \url{https://github.com/yunsungkim0908/analyticscore}.} as a reference framework. When applied to the domain of text-based constructed-response scoring, \textsc{AnalyticScore} outperforms many uninterpretable scoring methods in terms of scoring accuracy and is, on average, within 0.06 QWK of the uninterpretable SOTA across 10 items from the ASAP-SAS dataset. By comparing against human annotators conducting the same featurization task, we further demonstrate that the featurization behavior of \textsc{AnalyticScore} aligns well with that of humans.
\end{abstract}

\section{Introduction}

Accurate and credible assessment of knowledge and skills forms the basis for effective decision making in a variety of educational contexts, from student learning and instructional design to program development and policy making \cite{berman2019use}. 
When the set of knowledge and skills to be gauged involves complex, open-ended problem-solving and communication abilities, 
AI-driven 
\emph{automated scoring systems} can offer rapid, accessible, and scalable alternatives to the otherwise labor-intensive and costly process of training and deploying human scorers \cite{foltz2020past}.

Despite the growth in adoption and improvements in scoring capabilities, automated scoring of open-ended responses remains vulnerable to error, bias, and fairness concerns across scoring contexts \cite{hardy2026autoscoring,johnson2023evaluating}. 
To ensure that scoring decisions remain reliable, fair, and justifiable,
improving transparency and interpretability in automated scoring has become a moral imperative, not a mere technical preference \cite{khosravi2022explainable,holmes2022ethics,memarian2023fairness,schlippe2022explainability}.
% , especially in areas where accuracy still lags. 
In spite of the growing research on interpretable and explainable AI as well as its applications specifically within educational assessment, interpretable automated scoring remains mostly confined to academic research with limited adoption in large-scale, real-world assessment \cite{ies,whitmer2023results,whitmerlessons}.

% Ensuring the values of Fairness, Accountability, Transparency, and Ethics (FATE) in complex AI systems is now widely regarded as a moral imperative rather than a mere technical preference \cite{holmes2022ethics,memarian2023fairness}. In areas of education where AI increasingly plays a significant role, \emph{explainability} has emerged as an essential means of upholding the FATE principle \cite{khosravi2022explainable}. Especially in the high-stakes domain of educational assessment where untrustworthy and systematically flawed decisions could lead to profound negative impact on student equity, trajectories of learning, instructional decision-making, evaluations of educational programs/institutions, and broader public trust \cite{pellegrino2022learning,berman2019use}, the need for AI decision-making systems to be explainable is particularly more pronounced.

In this paper, we take a principled approach towards building a practical interpretable automated scoring solution for large-scale assessments. An effective interpretability solution begins by identifying the diverse needs of each stakeholder in understanding the system's decisions, and by grounding the development of interpretable AI systems in those needs \cite{bhatt2020explainable,preece2018stakeholders,paez2019pragmatic}. Research on interpretable automated scoring, on the other hand, has largely ignored this need-finding process. As we discuss in Section~\ref{sec:principles}, this neglect has often led to several claimed interpretability solutions that fail to address the diverse and nuanced interpretability needs of the human actors in educational assessment.

% This paper presents a key step towards building a practical solution by
% We hypothesize that the first step towards building a practical solution begins with
% establishing a set of explainability principles tailored to the needs of each assessment stakeholder.
% and developing a concrete, data-driven autoscoring framework grounded on these principles. 
% Previous works on explainability have shown that for an explainable AI system to lead to effective and wide adoption, its development should be grounded in a clear understanding of each stakeholder's unique requirements for comprehending the system and its decisions . 
% Research on explainable automated scoring has largely ignored this need-finding process, leading to several claimed solutions that fail to address the diverse and nuanced explainability needs (Section~\ref{sec:needs_and_principles}).

Grounded in the decades-long literature on educational assessment, we
% We
identify the needs and benefits of scoring model explanations for various large-scale assessment stakeholder groups consisting of test takers, assessment developers, and test users
% \footnote{\label{teachers}Teachers are also key stakeholders, but their interests often overlap with the other groups, so they are not discussed separately.}
(Section~\ref{sec:needs}). Targeted at those needs, we develop the principles of \textbf{faithful}, \textbf{grounded}, \textbf{traceable}, and \textbf{interchangeable} (\textbf{FGTI}) model interpretations for automated scoring (Section~\ref{sec:principles}). 

We further illustrate the feasibility of implementing the FGTI principles in practice and establish a concrete baseline for future research (Section~\ref{sec:analyticgrade}).  \textbf{\textsc{AnalyticScore}} is the first interpretable automated 
% short-answer 
scoring framework to embody our principles. It operates by extracting explicitly identifiable elements from unannotated student responses and featurizing each response into human-interpretable values based on those elements. 
These features are input to a logically traceable and human-interchangeable scoring module to produce a final scoring decision.
% These features are input to an intuitive ordinal logistic regression module for scoring.

We measure the performance of \textsc{AnalyticScore} on a real-world open-ended response dataset by measuring (1) scoring accuracy and (2) alignment of featurization behaviors with human judgments (Sections~\ref{sec:eval} and \ref{sec:results}).  \textsc{AnalyticScore} outperforms many uninterpretable scoring methods and is within 0.06 QWK of the uninterpretable SOTA on average across 10 items from the ASAP-SAS dataset. The featurization behavior of \textsc{AnalyticScore} also aligns well with humans (0.90, 0.72, 0.81 QWK across assessment areas). 

More broadly, this paper argues for a shift in how interpretable automated scoring should be formulated and studied: from treating automated scoring as an isolated prediction task to situating it within the broader stakeholder-centered assessment lifecycle spanning assessment design, scoring, and iterative refinement. 
Our findings indicate strong potential for implementing accurate and well-aligned solutions under this shift in perspective.

% To be able to produce a practical solution, the meaning and implementation of the abstract concept of ``explainability'' needs to be refined in the context of real-world problems and needs, as well as the intentions of the human actors involved \cite{lipton2018mythos,paez2019pragmatic}. 

% \paragraph{Automated Scoring and Interpretable AI}
\section{Related Work}

Automated scoring systems have been increasingly adopted across many assessment contexts over the past several decades, achieving acceptable levels of scoring accuracy in various areas of human learning \cite{whitmer2023results,whitmerlessons}. 
Despite progress, automated scoring of open-ended responses has yet to reliably obtain generalizable scoring accuracy across diverse scoring contexts \cite{hardy2026autoscoring}. The breadth of student problem-solving paths, language use, and fluency in expressing their reasoning makes consistent scoring particularly challenging across the full range of open-ended responses \cite{kim2023student}.
Even when automated scoring meets acceptable levels of scoring accuracy, errors or biases inherent in the scoring algorithm
can profoundly harm student learning and equity, policy evaluation, and public trust \cite{pellegrino2022learning,berman2019use}. 

% Automated essay scoring (AES) has evolved from early Bag-of-Words approaches and heavily engineered linguistic features to advanced neural methods that include LSTMs, CNNs, and transformer architectures. (Hardy, 2021) While feature-engineered systems have performed well, they often lack the ability to capture deeper semantic meaning, and transformer-based systems such as BERT require by-item tuning and loss of generalizable training. Generative language models (GLMs) like ChatGPT have attracted recent attention but raise significant sustainability, transparency, and cost concerns for widespread educational use. Consequently, smaller open-source models have emerged, though their in-context scoring generally lags behind fine-tuned classification approaches. Improved AES research thus calls for scalable self-attention strategies that handle entire essays without truncation, potentially through hierarchical or efficient attention mechanisms, while maintaining open-source accessibility to facilitate equitable use in classrooms and fostering truly explainable, domain-oriented assessment.
% \note{history of automated scoring and SOTA methods}

% As AI-driven automated scoring systems became more complex and opaque, researchers have increasingly noted the need to enhance the transparency of these systems through model explanations \cite{bennett2015validity,bauer2020cognitive,schlippe2022explainability}. 
% As AI-driven automated scoring systems became more complex and opaque, 
For these reasons, researchers have increasingly noted the need to enhance the transparency of complex AI-driven automated scoring systems through model explanations \cite{bennett2015validity,bauer2020cognitive,schlippe2022explainability}. 
Several approaches have been proposed to address this challenge, and we discuss them in detail in Section~\ref{sec:principles} in connection with our four principles.
% include using LLMs to provide scoring rationale alongside scoring decisions \cite{li2025automated,li2023distilling,lee2024applying}, calculating feature importance values \cite{kumar2021automated,kumar2020explainable,schlippe2022explainability,asazuma2023take}, displaying feature attribution maps \cite{schlippe2022explainability,li2025automated}, presenting confidence metrics for scoring decisions \cite{conijn2023effects} \todo. As we discuss in Section~\ref{sec:principles}, these existing methods do not fulfill one or more of the principles of grounded, faithful, traceable and correctable explanations, which are crucial to addressing the needs of the human actors in the assessment process. 

Despite growing research interests, interpretable automated scoring still lacks practical adoption and meaningful real-world use. The 2023 NAEP Math Automated Scoring Challenge\footnote{\url{https://github.com/NAEP-AS-Challenge/math-prediction}} for open-ended math responses organized by the US National Center for Education Statistics (NCES) found that none of the submissions met the criteria for interpretability despite several methods achieving near-human scoring accuracy \cite{ies,whitmerlessons}. Our work provides a stakeholder-centered approach towards addressing this gap.

\section{Building the Principles of Interpretable Automated Scoring}
\label{sec:needs_and_principles}

% This section identifies the different human actors in the learning assessment ecosystem and their needs and potential benefits for explainable automated scoring. We then motivate four foundational design principles of explainable automated scoring methods to address these needs and benefits.

Insights derived from scoring support various human actors throughout the overall assessment process. 
% Different human actors in the learning assessment ecosystem have unique needs and potential benefits from explainable automated scoring. 
Below we analyze three main stakeholder groups in large-scale assessment -- test takers, assessment developers, and test users \cite{berman2019use,standards}. Each stakeholder group's distinct roles and priorities uniquely shape how interpretable automated scoring can improve their assessment experience.

\subsection{Interpretability Needs and Benefits}
\label{sec:needs}
\paragraph{Test Takers} 
% Students' needs and benefits for interpretable scoring differ for summative and formative assessments. \emph{Summative assessments} are assessments \emph{of} learning 
% such as standardized tests, college entrance exams, or final exams 
% that evaluate student achievement, assign grades, or determine proficiency levels after learning took place \cite{harlen2005teachers}. The results of these assessments are often used to make high-stakes decisions about students, which makes it important to ensure that students can trust the justifiability and fairness of the scoring decisions made on their behalf \cite{williamson2012framework}. 

The needs and benefits of interpretable scoring vary depending on the assessment type: summative or formative. Most large-scale assessments are summative assessments, which are assessments \textit{of} learning that support evaluating learner achievement, assigning grades, or determining proficiency levels \cite{harlen2005teachers}. Because these assessments often drive high-stakes decisions, test takers need to trust the fairness and justifiability of scoring decisions \cite{williamson2012framework}. 
% In \emph{summative assessments}, where results often influence high-stakes decisions about students, it is important that students trust the fairness and justifiability of scoring decisions \cite{williamson2012framework}.
Provided that the scoring algorithm implements sound scoring logic, allowing test takers or their representatives to examine traceable explanations for scoring decisions can foster trust \cite{bauer2020cognitive,ferrara2022validity}. 
% If the explanations can further facilitate an efficient process for identifying and justifiably correcting erroneous scoring results, they can also assist in enforcing a simple and streamlined quality control process to flag and remedy potential scoring errors 
These explanations can also support a streamlined quality control process by facilitating the identification and correction of errors, improving the overall integrity of the assessment 
(see \citet{bennett2015validity} and \citet{ferrara2022validity}).
% Given also that several states and major testing agencies have procedures for challenging and appealing scores (\citealt{neill1997testing}, see also Massachusetts 603 CMR \textsection 30.08 ``Score Appeals''), 

Formative assessments are assessments \textit{for} learning, intended to guide and improve learner performance through frequent practice, progress monitoring, and skill diagnosis \cite{wiliam2011embedded,black1998assessment}. In this context, 
% For \emph{formative assessments}, 
the function of automated scoring is primarily to provide timely, effective and actionable feedback to support learning \cite{bennett2006moving,dicerbo2020assessment}. Effective feedback should help learners understand the discrepancy between their work and a desired outcome  \cite{schwartz2016abcs}. A step-by-step explanation of the features observed in a learner’s work, coupled with human-understandable descriptions of how those features were processed can be used to provide such elaborative feedback.
% specific and elaborative
% Explanations about the features extracted, along with the internal reasoning structure of the scoring model can be used to give feedback that is both specific and elaborative. If these explanations can further facilitate the identification of what students \emph{should have} done to produce desired outcomes, the resulting feedback will better enable students to adjust and improve future actions.

\paragraph{Assessment Developers}
% The reliability of scoring algorithms in identifying evidence of the constructs (target knowledge, skills, and abilities) measured by the task is critical \cite{bejar2016automated}.
Scoring algorithms should reliably identify evidence of the constructs (target knowledge, skills, and abilities) measured by the task \cite{bejar2016automated}.
Understanding the types of evidence that an automated scoring algorithm reliably detects also informs other key aspects of assessment design, such as 
% the selection of constructs to measure and the design of tasks that effectively elicit such evidence 
construct selection and task design \cite{bennett1998validity}. Model explanations can facilitate this understanding by transparently revealing the features used by the scoring algorithm and its intermediate reasoning steps. Explanations can also help determine which parts of the algorithm can be reused, avoiding the costly and time-consuming process of training a new scoring algorithm for each new task (see \citet{dicerbo2020assessment}).

Model explanations also yield specific insights into areas where the scoring model can be improved and how. Scoring models often need to be tuned for various reasons. For instance, models trained on data may reflect biases related to response strategies specific to student groups  \cite{ferrara2022validity,rupp2018designing}. Scoring models may also become less stable over time as the test-taker population and/or scoring criteria change  \cite{bejar2016automated}. Transparent inspection of model decisions helps identify problematic model elements, enabling targeted data collection and modified training objectives to improve the model. 

\paragraph{Test Users} Test users, including professionals who select and administer tests, educators, administrators, and policymakers, depend on score reliability and interpretation validity to make system-level or instructional decisions. Their reliance on the integrity and validity of scores to drive decisions is significant \cite{standards}. Model explanations provide concrete evidence to validate the choice of the scoring model\footnote{More examples of validity arguments on the use of automated scoring can be found in \cite[Table 7.7]{bennett2015validity}.}. This includes understanding whether the extracted features and scoring logic fully capture the rubric and the construct definition, and whether the internal structure of the automated scores align with the construct of interest \cite{bennett2015validity}.

\subsection{The FGTI Interpretability Principles}
\label{sec:principles}

% \begin{figure}[t]
%     \centering
%     \includegraphics[width=\linewidth]{img/principles.pdf}
%     \caption{Caption}
%     \label{fig:enter-label}
% \end{figure}

We develop four foundational interpretability principles -- \textbf{F}aithful, \textbf{G}rounded, \textbf{T}raceable, and \textbf{I}nterchangeable (\textbf{FGTI}) -- targeting the needs and benefits of large-scale assessment stakeholders from Section~\ref{sec:needs}. Our first foundational principle 
% for supporting the aforementioned needs and potential benefits of the assessment stakeholders 
is that explanations should be \emph{faithful} \cite{jacovi2020towards}. Faithfulness is an important requirement in many high-stakes applications of interpretable AI \cite{rudin2019stop}. Similar expectations extend to assessments, and all of the needs and benefits outlined in Section~\ref{sec:needs} depend crucially on faithfulness.
\begin{principle}[Faithful]
    \label{prin:faithful}
    Explanations of scoring decisions should accurately reflect the computational mechanism behind the scoring model's prediction.
\end{principle}

A notable example of \textbf{un}faithful scoring explanations are texts produced by prompting LLMs to generate an explanation (e.g., \citet{lee2024applying} and \citet{li2025automated}). Stepwise reasoning verbalized by LLM through prompting strategies such as chain-of-thought \cite{wei2022chain} are not explanations of their internal computation \cite{sarkar2024large} and often fail to reflect the model's true reasoning behavior \cite{turpin2023language,arcuschin2025chain}. Moreover, LLMs are highly sensitive to superficial changes in prompts and input text, frequently exhibiting inconsistent judgments \cite{wang2024large}. Therefore, even when LLMs achieve high scoring accuracy, prompting them to ``explain their decisions'' cannot reliably address the stakeholder needs identified in Section~\ref{sec:needs}.

Next, the model should use meaningful features that are explicitly linked to each student's work and rely only on those features for the downstream computation.
\begin{principle}[Grounded]
    \label{prin:grounded}
    Initial features computed by the scoring model should represent human-understandable, explicitly identifiable elements of student work and item task.
\end{principle}
Regardless of the routine used to derive those features, these feature values should possess meaning that is understandable to humans and be explicitly grounded in both the student work and the item/task. 
For instance, cosine similarity of sentence embeddings used as an input feature (e.g., \citet{condor2024explainable}) is less human-understandable than discrete features whose values are associated with clear, verbalizable meaning. Using grounded features allows the evidence used by the scoring engine to be scrutinized.

% How the model processes these features should be either computationally or logically traceable:
% \begin{principle}[Traceable]
%     \label{prin:traceable}
%     The mapping from features to the final score must either (1) be simulatable\footnote{Per \cite{lipton2018mythos}, a \emph{simulatable} model is one whose entire calculation can be done by a human in a reasonable timeframe.}, or (2) be composed of modules that each evaluate specific human-understandable properties from clearly specified inputs and collectively account for the model's entire logic.
% \end{principle}
% We refer to (1) as \textbf{computationally traceable} and (2) as \textbf{logically traceable}. Computational traceability is both feasible and appropriate for simple tasks, especially when the model's logic is accurate and aligned with the intended score assignment. Even when additional inference is required beyond explicitly identifiable features (Principle~\ref{prin:grounded}), not all features need to be interpretable to allow for logical traceability -- a requirement that is rarely feasible in modern ML models. The interpretable features, however, should be specific and account for the overall scoring logic without leaving any unexplained dependencies. 

\begin{figure*}[t!]
    \centering
    \includegraphics[width=\linewidth]{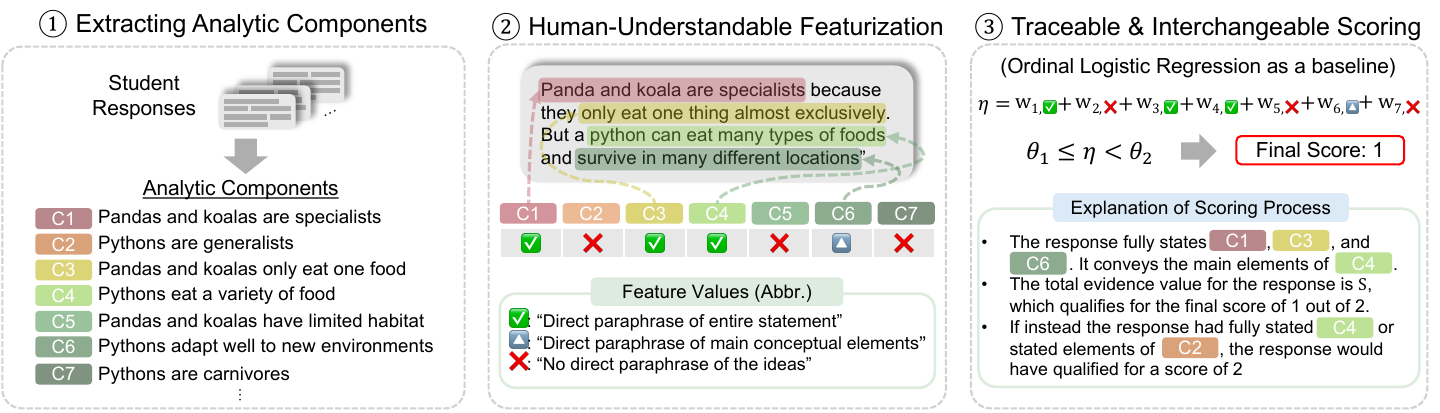}
    \caption{Schematic of the \textsc{AnalyticScore} framework. The example question is: \textbf{``Explain how pandas in China and koalas in Australia are similar, and how they both are different from pythons.''}}
    \label{fig:analyticgrade}
\end{figure*}

How should the model process these features to ultimately produce a final score? Scoring is inherently an evidentiary reasoning process, where elements of the student response and item tasks serve as evidence to support the inference about student knowledge and skills that the score represents \cite{dicerbo2020assessment,mislevy2020evidentiary}. Stakeholders need to be able to inspect and interact with the internal structure of the scoring model to ensure soundness, construct-relevance, and fairness (Section~\ref{sec:needs}). To meet this need, the model’s evidentiary reasoning process must be decomposable into clear, sequential steps that a human could reliably execute and possibly intervene. Our next 2 principles state that the scoring model should be conducive to this decomposition and intervention:
\begin{principle}[Traceable]
    The scoring model should consist of subroutines that each represent a specific, well-defined evidentiary reasoning step on clearly specified inputs. 
    \label{prin:traceable}
\end{principle}
\begin{principle}[Interchangeable]
    A human should be able to act interchangeably on each of the reasoning subroutines.
    \label{prin:interchangeable}
\end{principle}
Not all intermediate representations calculated by the model need to be understandable to humans, but the reasoning subroutines should collectively account for the entire scoring logic. Moreover, humans should be able to act interchangeably with each subroutine and replace its outputs with human-generated results if deemed necessary.

Many proposed interpretability approaches are not grounded, traceable, or interchangeable in the sense described above. These include, for instance, calculating feature importance values \cite{kumar2021automated,kumar2020explainable,schlippe2022explainability,asazuma2023take}, displaying feature attribution maps \cite{schlippe2022explainability,li2025automated}, or presenting confidence metrics for scoring decisions \cite{conijn2023effects}. This limits the capacity to thoroughly inspect the model’s features and internal decision-making structure.

\section{A Principled Framework for Interpretable Automated Scoring}
\label{sec:analyticgrade}

To illustrate the feasibility of implementing the FGTI principles and to set a baseline for future research, we present \textsc{AnalyticScore} as a reference framework and demonstrate it in the domain of text-based constructed-response scoring. In this setting, students write a short one- to five-sentence answer in response to an assessment item which is scored with an emphasis on content correctness and demonstrated reasoning  
% (e.g., not grammar) 
\cite{leacock2003c,shermis2015contrasting}. The scoring model has access to a training set of student response texts paired with human-annotated scores $(r_1,s_1),...,(r_n,s_n)$ as well as unannotated responses $\{r_{n+1},...,r_m\}$. The goal at inference time is to predict the score $s$ for a new response $r$.
% Compared to automated essay scoring which focuses on the overall quality of writing and is already operational at scale for many real-world assessments, automated short-answer scoring remains an active area of research. \todo

\textsc{AnalyticScore} (Figure~\ref{fig:analyticgrade}) is a modular, 3-phase framework embodying the FGTI principles. Phase 1 identifies explicitly grounded \textit{analytic components} to be used. Phase 2 catalogs, or \textit{featurizes}, the presence of these components in student responses. Phase 3 uses the features to compute a score. Phases 1 and 2 depend only on the response texts without any annotations. Human score labels are only used during Phase 3. 
% We articulate each phase of the framework below.

\subsection{Phase 1: Extracting Analytic Components} With the student responses in the training set (and optionally the content of the assessment item), \textsc{AnalyticScore} first extracts a set of \textbf{analytic components}, which are explicitly identifiable elements of student responses described in Principle~\ref{prin:grounded}. 
% In this work, we consider a specific type of components which are representative, atomic units of explicit statements, arguments, or claims as in Figure~\ref{fig:analyticgrade}.
% Components are expressed in natural language in alignment with Principle~\ref{prin:interchangeable} and are \textit{propositions} \cite{graesser_constructing_1994}, explicit atomic units of inference, such as simple statements or claims
\[
    [c_1, ..., c_k] = \text{Extract}(r_1,...,r_m),
\]
For the scoring task considered in this work, we operationalize analytic components as propositions.

Using student responses to extract analytic components aligns with the broader workflow of assessment development and refinement. During these steps, assessment developers continuously analyze raw student responses from pilot tests and field tests to create and iteratively refine scoring rubrics, scoring directions, and reference responses \cite{mccaffrey2022best}. The analytic component extraction phase of \textsc{AnalyticScore} can be incorporated into this workflow by generating candidate analytic components from raw student responses, which assessment developers can also review and refine. 

Provided the extracted components align with the intended item design and are human understandable, any procedure can be used for component extraction. 
% If necessary, these components could also be manually reviewed and modified by assessment designers. 
In this work, we implement component extraction by prompting an LLM with the prompts shown in Appendix~\ref{sec:appendix_prompt}. 
Having too many analytic components could reduce interchangeability (Principle~\ref{prin:interchangeable}) by increasing the number of processed features \cite{lipton2018mythos}. We therefore limit extraction to 15 components per scoring unit.

% Although analytic components may appear similar to ``rubrics'' used in scoring, there are key differences. Rubrics are  reproducible with the prompts shown in Figure~\ref{fig:prompts}

% Moreover, not all 

\subsection{Phase 2: Featurizing Responses}
\label{sec:featurize}
Once the analytic components have been identified, student responses are featurized according to the presence of these components $c_1,...,c_k$ in each response $r$. This step uses a custom-defined labeling function $f(r;c)$ whose outputs are associated with human-understandable meaning (Principles ~\ref{prin:grounded} \& ~\ref{prin:traceable}). For this study, we defined the following general purpose labeling function designed for constructed-response scoring. The precise definition of each score category can be found in Appendix~\ref{sec:appendix_prompt}:
\begin{equation*}
    f(r;c) = \left\{\begin{array}{l}
        2, \text{ $r$ contains direct paraphrase of $c$} \\
        1, \text{ $r$ contains partial paraphrase of $c$} \\
        0, \text{ $r$ doesn't contain paraphrase of $c$}
    \end{array}\right.
\end{equation*}

We implement $f(r;c)$ with Chain-of-Thought prompting \cite{wei2022chain} using the prompts shown in Appendix~\ref{sec:appendix_prompt}. (Note that CoT is used solely as a prompting technique, and the generated ``thoughts'' are explicitly discarded.) Inspired by the self-consistency decoding strategy for LLMs \cite{wang2022self}, we apply the first-to-three aggregation rule to consider the possibly diverse interpretations of the labeling criteria when selecting the final output. The one-hot encodings of each $f(r;c)$ are then concatenated to produce a $3k$-dimensional binary featurization of $r$:
\begin{equation*}    
    F(r) = 
    \text{OneHot}(f(r;c_1))\concat\cdots \concat \text{OneHot}(f(r;c_k))
\end{equation*}

\paragraph{Distilling LLM Featurizer into Open Source}

Using proprietary LLMs for featurization can quickly become too expensive in large-scale assessment settings, especially with many responses to score and analytic components to consider. To avoid the linearly growing cost of featurization, a small open-source model can be supervised fine-tuned using a subsample of $(r,c)$ pairs, where $r$ is a response from the training set and $c$ is an analytic component from Phase 1. In this paper, we randomly sampled 10k pairs across all 10 items, calculated the featurization labels on these samples using o4-mini, and collected the LLM model requests and outputs generated during this process that aligned with the final aggregated label. This dataset was used to fine-tune Llama-3.1-8b-instruct with QLoRA~\cite{dettmers2023qlora}.

\subsection{Phase 3: Logically Traceable Scoring}
\label{sec:scoring}

Based on the featurized responses, a traceable and interchangeable scoring model (Principles~\ref{prin:traceable} and \ref{prin:interchangeable}) is trained using the labeled response pairs $(r_1,s_1),...,(r_n,s_n)$. While any traceable and interchangeable model can be used, given the ordinal nature of the score categories, we instantiate the scoring module as the Immediate-Threshold variant of Ordinal Logistic Regression \cite{rennie2005loss,pedregosa2017consistency} in this work. 
% As with other phases of \textsc{AnalyticScore}, the scoring module can be chosen to be any model that is traceable and interchangeable. In this work, given the nature of the score categories, we demonstrate it using the Immediate-Threshold variant of Ordinal Logistic Regression \cite{rennie2005loss,pedregosa2017consistency} as our scoring module. 
Combined with the one-hot encoding featurization from Phase 2, the resulting algorithm calculates the sum of weights for each component and feature label: $\eta = \sum^c_{i=1} w_{i,f(r,c_i)}$, where $w$ are the trained weights. Scores are determined by comparing $\eta$ to a set of learned thresholds $\theta_j$; the predicted score corresponds to the ordinal category $j$ for which $\theta_j \leq \eta < \theta_{j+1}$. $\eta$ can be understood as an ``evidence value'' used for scoring.

\subsection{Interpretability of \textsc{AnalyticScore}}

An example of \textsc{AnalyticScore}'s model explanation is shown in the far right panel of Figure~\ref{fig:analyticgrade}. By demonstrating human-understandable features of the response (Principle~\ref{prin:grounded}) and the exact decision process (Principle~\ref{prin:traceable}), the explanation transparently and faithfully reveals the actual scoring mechanism used (Principle~\ref{prin:faithful}). If, based on the explanation, the model is suspected to have made an error (e.g., C6 should be a check, not a triangle), a human inspector can modify the featurization and recalculate the score by following the scoring algorithm (Principle~\ref{prin:interchangeable}), which is also how the ``if instead...'' explanation is generated.

The structure of \textsc{AnalyticScore}'s scoring model is akin to Concept Bottleneck Models \cite{koh2020concept,yang2023language} in that we enforce a layer of intermediate representations with human-understandable ``concepts.'' Our framework ensures that the intermediate features have human-understandable values associated with explicitly identifiable elements of student work (Principle~\ref{prin:grounded}), as opposed to inferred characteristics.

\subsection{Generalizing to Other Response Types}

In this paper, we illustrate \textsc{AnalyticScore} in the specific context of text-based constructed-response scoring. It is important to note, however, that \textsc{AnalyticScore} is a generalized framework whose concept can be applied to other response types. For instance, in programming, analytic components could represent recognizable coding patterns (e.g., for-loop/while-loop, recursion), features of the decomposition structure, or identifiable characteristics of program outputs. For game- or simulation-based assessments, analytic components could reflect elements of the produced artifacts (e.g., written justifications, diagrams, or plans) or interpretable patterns in the test-taker's action sequence. An important direction for future work is to study how each phase of \textsc{AnalyticScore} (component extraction, featurization, and traceable \& interchangeable scoring) should be adapted for different response modalities.

\section{Evaluating \textsc{AnalyticScore}}
\label{sec:eval}
We now evaluate \textsc{AnalyticScore}’s scoring accuracy and how well its featurization aligns with human judgments on a real-world scoring dataset.
% Having introduced \textsc{AnalyticScore} and discussed its interpretability, we now evaluate its scoring performance and how its featurization aligns with human judgments on a real-world dataset.

\paragraph{Dataset} The ASAP-SAS dataset \cite{shermis2015contrasting}\footnote{\url{https://www.kaggle.com/competitions/asap-sas/data}} is the largest publicly available dataset of English constructed-responses from US schoolchildren for 10 different open-ended exam questions, covering 3 areas of assessment: Science, Reading (Informational Text), and Reading (Literature Text).
% four of which were originally handwritten and transcribed
Human raters double-scored and assigned a single number to each student response using a 3 or 4 point rubric. The dataset details are reported in Appendix~\ref{sec:appendix_asap}. We use the original test set and split the public training set into training and validation sets with an 8:2 ratio.

\paragraph{\textsc{AnalyticScore} Implementation Details} For each assessment item, we used \texttt{GPT-4.1} as the base LLM and extracted 15 analytic components except for Q7. This item uses a two-part scoring scheme to separately assess a character trait identified from the reading and its supporting evidence. We extracted 15 analytic components from each part, totaling 30 components. For the featurizer, we experimented with GPT-4.1-mini and Llama-3.1-8B-Instruct as our base LLM, each with temperature set to 0.7 and 1.0. Training and implementation details can be found in Appendix~\ref{sec:appendix_implementation_details}.
% With GPT-4.1-mini, in addition to first-to-3 aggregation we also implemented the \textbf{deterministic} variant by setting temperature to 0 and making a single call. 

{
\setlength{\tabcolsep}{0.7mm}

\begin{table*}[t]

\small
\centering
\begin{tabular}{lcccccccccccccc}
\toprule
 & \textbf{Q1} & \textbf{Q2} & \textbf{Q3} & \textbf{Q4} & \textbf{Q5} & \textbf{Q6} & \textbf{Q7} & \textbf{Q8} & \textbf{Q9} & \textbf{Q10} & \textbf{All Avg.} & \textbf{Sci Avg.} & \textbf{R(Inf) Avg.} & \textbf{R(Lit) Avg.} \\
\midrule
\textbf{Human Scorers} & 0.95 & 0.93 & 0.77 & 0.75 & 0.95 & 0.93 & 0.96 & 0.86 & 0.84 & 0.87 & 0.88$\pm$0.02 & 0.93$\pm$0.01 & 0.79$\pm$0.03 & 0.91$\pm$0.05 \\
\multicolumn{14}{l}{\textbf{\textsc{AnalyticScore}}}  &  \\
% \quad w/ GPT-4.1-mini & 0.80 & 0.85 & 0.58 & 0.63 & 0.82 & 0.77 & 0.60 & 0.62 & 0.80 & 0.68 & 0.72$\pm$0.03 & 0.78$\pm$0.03 & 0.67$\pm$0.07 & 0.61$\pm$0.01 \\
% noinst & 0.81 & 0.84 & 0.54 & 0.65 & 0.80 & 0.80 & 0.59 & 0.53 & 0.77 & 0.66 & 0.70$\pm$0.04 & 0.78$\pm$0.03 & 0.66$\pm$0.07 & 0.56$\pm$0.03 \\
% 2way & 0.82 & 0.83 & 0.65 & 0.42 & 0.75 & 0.72 & 0.57 & 0.56 & 0.75 & 0.60 & 0.67$\pm$0.04 & 0.75$\pm$0.04 & 0.61$\pm$0.10 & 0.57$\pm$0.00 \\
% nocot & 0.82 & 0.85 & 0.63 & 0.64 & 0.80 & 0.80 & 0.57 & 0.51 & 0.82 & 0.70 & 0.71$\pm$0.04 & 0.79$\pm$0.03 & 0.70$\pm$0.06 & 0.54$\pm$0.03 \\
\quad w\!/ GPT-4.1-mini\textsuperscript{*} & 0.80 & \textbf{0.86} & 0.64 & 0.59 & 0.79 & 0.78 & 0.61 & 0.59 & 0.80 & 0.68 & 0.72$\pm$0.03 & 0.78$\pm$0.03 & 0.68$\pm$0.06 & 0.60$\pm$0.01 \\
\quad w\!/ Llama-3.1-8b (I)\textsuperscript{*} & 0.57 & 0.57 & 0.59 & 0.56 & 0.69 & 0.47 & 0.52 & 0.45 & 0.74 & 0.60 & 0.58$\pm$0.03 & 0.58$\pm$0.04 & 0.63$\pm$0.05 & 0.48$\pm$0.04 \\
\quad\quad + Distillation\textsuperscript{*} & 0.80 & \underline{0.82} & 0.68 & 0.59 & 0.81 & 0.76 & 0.62 & 0.59 & 0.78 & 0.64 & 0.71$\pm$0.03 & 0.77$\pm$0.03 & 0.68$\pm$0.06 & 0.60$\pm$0.01 \\
\multicolumn{14}{l}{\textbf{Few-Shot}}  &  \\
\quad GPT-4.1 & 0.69 & 0.65 & 0.61 & 0.65 & 0.72 & 0.61 & 0.34 & 0.57 & 0.76 & 0.69 & 0.63$\pm$0.04 & 0.67$\pm$0.02 & 0.68$\pm$0.04 & 0.45$\pm$0.12 \\
\multicolumn{14}{l}{\textbf{Supervised LLM}}  &  \\
\quad BERT & 0.80 & 0.80 & \underline{0.70} & 0.70 & 0.80 & 0.81 & 0.69 & \textbf{0.68} & \textbf{\textit{0.84}} & 0.71 & 0.75$\pm$0.02 & 0.79$\pm$0.02 & 0.74$\pm$0.05 & \textbf{0.69$\pm$0.01} \\
\quad DeBERTa & \underline{0.85} & \textbf{0.86} & 0.66 & 0.70 & 0.81 & \underline{0.83} & \underline{0.71} & 0.64 & 0.79 & 0.71 & \underline{0.76$\pm$0.03} & \underline{0.81$\pm$0.03} & 0.72$\pm$0.04 & 0.67$\pm$0.04 \\
\quad Llama-3.1-8b (I) & 0.84 & 0.73 & \textbf{0.72} & \underline{0.71} & \underline{0.82} & 0.81 & \underline{0.71} & \underline{0.66} & \underline{0.82} & \underline{0.75} & \underline{0.76$\pm$0.02} & 0.79$\pm$0.02 & \underline{0.75$\pm$0.03} & \textbf{0.69$\pm$0.02} \\
\quad\quad w/ rubric & \textbf{0.87} & 0.80 & 0.68 & \textbf{\textit{0.77}} & \textbf{0.85} & 0.80 & \textbf{0.72} & 0.65 & \textbf{\textit{0.84}} & \textbf{0.79} & \textbf{0.78$\pm$0.02} & \textbf{0.82$\pm$0.02} & \textbf{0.76$\pm$0.05} & \underline{0.68$\pm$0.04} \\
\quad Llama-3.1-8b  & 0.83 & 0.75 & \underline{0.70} & \textbf{\textit{0.77}} & \underline{0.82} & \textbf{0.84} & 0.68 & 0.65 & \underline{0.82} & 0.74 & \underline{0.76$\pm$0.02} & 0.80$\pm$0.02 & \textbf{0.76$\pm$0.03} & 0.67$\pm$0.02 \\
\multicolumn{14}{l}{\textbf{Baseline}}  &  \\
\quad AutoSAS & 0.68 & 0.47 & 0.57 & 0.61 & 0.50 & 0.54 & 0.37 & 0.44 & 0.77 & 0.68 & 0.56$\pm$0.04 & 0.57$\pm$0.04 & 0.65$\pm$0.06 & 0.41$\pm$0.04 \\
\quad ASRRN & 0.60 & 0.43 & 0.57 & 0.60 & 0.61 & 0.64 & 0.59 & 0.51 & 0.71 & 0.66 & 0.59$\pm$0.02 & 0.59$\pm$0.04 & 0.63$\pm$0.04 & 0.55$\pm$0.04 \\
\quad NAM\textsuperscript{*} & 0.63 & 0.62 & 0.43 & 0.35 & 0.72 & 0.63 & 0.42 & 0.38 & 0.76 & 0.62 & 0.56$\pm$0.05 & 0.64$\pm$0.02 & 0.52$\pm$0.13 & 0.40$\pm$0.02 \\
\bottomrule
\end{tabular}

\caption{Test-time Quadratic Weighted Kappa (QWK) of scoring models per item, along with average per assessment area. \textbf{Best}, \underline{second-best}, and \textbf{\textit{human-level}} performance scores are marked respectively.  \textbf{Sci.}: Science (Q1,2,5,6). \textbf{R(Inf)}: Reading (Informational Text) (Q3,4,9). \textbf{R(Lit)}: Reading (Literature) (Q7,8). \textbf{(*)} are methods that are considered interpretable. Human scoring accuracy was taken from \cite{shermis2015contrasting}.}
\label{tab:results}

\end{table*}
}

\subsection{Scoring Accuracy Experiment}

We measured scoring accuracy in terms of quadratic weighted kappa (QWK) against the reference scores in the test set, following the convention of the automated scoring literature \cite{shermis2015contrasting,ies}. 
%QWK is defined as $\kappa = \sum\sum w_{ij}x_{ij} / \sum\sum w_{ij}x_{ij}(1-x_{ij})$ where the weights $w_{ij}$ increase quadratically with rater discrepancy from observed probability $x_{ij}$. 
We compare against the following baseline methods (see Appendix~\ref{sec:appendix_implementation_details} for the list of hyperparameters).:
% \footnote{Hyperparameter values used are shown in the appendix.}
\begin{description}

  \item[Few-Shot Prompting:] We few-shot prompt \textbf{GPT-4.1} with 10 randomly selected responses from each score category, including a rubric for the score categories.

  \item[Supervised Fine-tuned LLM:] The following LLM-based classifiers were fine-tuned on the response-score pairs: \textbf{BERT}~\cite{devlin2019bert}, \textbf{DeBERTa}~\cite{he2020deberta}, \textbf{Llama-3.1-8b}, and \textbf{Llama-3.1-8b-Instruct}~\cite{grattafiori2024llama}. We also fine-tune \textbf{Llama-3.1-8B-Instruct} with the scoring rubric added as a part of the input.

  \item[Automated Scorer Baselines:] \textbf{AutoSAS}~\cite{kumar2019get}, \textbf{AsRRN}~\cite{li2023answer}, and \textbf{NAM}~\cite{condor2024explainable}.

\end{description}
The only baseline method that has aspects of interpretability is NAM. This method requires hand-crafting a specific form of rubric describing the key phrases and concepts to be used by the response. Using sentence embeddings with n-gram matching as its features, this method implements a logistic regression score classifier. To implement this baseline, we replace the rubrics with the analytic components extracted by our \textsc{AnalyticScore}. 
% We attain the same level of results for the subset of the ASAP data reported in the NAM paper and then extend the model to the left out items.

% Although the use of sentence embeddings as features makes the algorithm less grounded (Principle~\ref{prin:grounded}), we will consider this an interpretable baseline.

% \textbf{Few-Shot Prompting}: We few-shot prompt \textbf{GPT-4.1} with 10 randomly selected responses from each score category, including a qulitative rubric of the score categories.

% \textbf{Supervised Fine-tuned LLM}: We fine-tune an LLM-based classifier on the training response-score pairs. We explore the following range of models: \textbf{BERT}~\cite{devlin2019bert}, \textbf{DeBERTa}~\cite{he2020deberta}, \textbf{Llama-3.1-8b}, and \textbf{Llama-3.1-8b}~\cite{grattafiori2024llama}. We also fine-tune Llama-3.1-8B-Instruct with a qualitative rubric of the score categories added to the input.

% \textbf{Automated Scorer Baselines}: We also implemented various notable automated scoring methods including \textbf{AutoSAS}~\cite{kumar2019get}, \textbf{AsRRN}~\cite{li2023answer}, and \textbf{NAM}~\cite{condor2024explainable}.

\subsection{Featurization Alignment Experiment}
\label{sec:eval_alignment}

The feature labeling task described in Figure~\ref{fig:prompts} was designed to produce human-\emph{understandable} features (Principle~\ref{prin:grounded}). But how well does the LLM’s featurization behavior align with how humans actually understand this task? Even more fundamentally, how well do humans themselves agree in their understanding of this task?

To answer these questions, we sampled 50 (response, analytic component) pairs for each of the 3 assessment areas. To ensure balanced representation, the sample included a balanced number of pairs from each of the three score categories, as initially determined by the GPT-4.1-mini featurizer. We then asked 7 human annotators to conduct the labeling task on these samples. The human annotators consisted of five volunteers from an R1 institution and two of the study's authors. None of the annotators had prior exposure to the LLM's featurization outputs, preventing any potential bias. All annotators had advanced academic training (PhD-level) and teaching experience, five of whom have been instructors at the primary, secondary, and/or post-secondary level. Additional details about the annotation process can be found in Appendix~\ref{sec:appendix_alignment_study}.

Human label was generated by majority voting (ties resolved randomly). We calculated inter-rater reliability among human labelers (Krippendorff's $\alpha$) and alignment between each LLM featurizer and human labels (QWK and class-wise F1). We report the 95\% bootstrap CI of each metric, reweighting the sampling probability to account for the initial balanced sampling of score categories.

\section{Experiment Results}
\label{sec:results}

{
\setlength{\tabcolsep}{1mm}
\begin{table*}[t]
\centering
\small

\begin{tabular}{clcccccccc}
\toprule
% \textbf{Task Type} & \multicolumn{1}{c}{\textbf{Model}} & \textbf{QWK} & \textbf{F1-Macro} & \textbf{F1 (Class 2)} & \textbf{F1 (Class 1)} & \textbf{F1 (Class 0)} \\
\multirow{2}{*}{\textbf{Assessment Area}} 
& \multirow{2}{*}{\textbf{Featurizer Model}} 
& \multirow{2}{*}{\textbf{QWK}} 
% & \multicolumn{3}{c}{\textbf{(Proj.) Label Distribution}}
& \multicolumn{3}{c}{\textbf{Label Distribution}\footref{distribution}}
& \multicolumn{3}{c}{\textbf{Label-wise F1}}
\\
\cmidrule(lr){4-6}
\cmidrule(lr){7-9}
& & & 2 & 1 & 0 & 2 & 1 & 0 \\
\midrule

% \multirow{6}{*}{\makecell{Expository\\(Scientific/Empirical)}} 
% & \multicolumn{6}{c}{Krippendorff's $\alpha$ among Human Raters: $(0.718, 0.723)$} \vspace{1ex}
% \\
\multirow{4}{*}{Science} 
 & \textbf{\underline{Human}} & & 15.32\% & 3.70\% & 80.98\% & & & \\
 & GPT-4.1-mini & (0.89, 0.89) & 7.56\% & 12.59\% & 79.85\% & (0.83, 0.84) & (0.20, 0.21) & (0.96, 0.96) \\
 & o4-mini & (0.94, 0.95) & 14.35\% & 8.17\% & 77.47\% & (0.93, 0.93) & (0.49, 0.51) & (0.98, 0.98) \\
 & Llama-3.1-8B (Distilled) & (0.90, 0.90) & 11.54\% & 5.27\% & 83.19\% & (0.89, 0.89) & (0.20, 0.22) & (0.97, 0.97) \\

\midrule
\multirow{4}{*}{\makecell{Reading\\(Informational\\Text)}} 
 & \textbf{\underline{Human}} & & 11.48\% & 18.44\% & 70.08\% & & & \\
 & GPT-4.1-mini & (0.71, 0.72) & 9.70\% & 22.01\% & 68.29\% & (0.68, 0.69) & (0.54, 0.55) & (0.87, 0.88) \\
 & o4-mini & (0.81, 0.82) & 18.05\% & 16.54\% & 65.42\% & (0.73, 0.74) & (0.69, 0.70) & (0.94, 0.94) \\
 & Llama-3.1-8B (Distilled) & (0.72, 0.73) & 20.20\% & 9.98\% & 69.82\% & (0.61, 0.62) & (0.24, 0.26) & (0.91, 0.91) \\

\midrule
\multirow{4}{*}{\makecell{Reading\\(Literature)}} 
 & \textbf{\underline{Human}} & & 9.33\% & 5.12\% & 85.55\% & & & \\
 & GPT-4.1-mini & (0.52, 0.54) & 2.64\% & 10.67\% & 86.69\% & (0.44, 0.46) & (0.67, 0.69) & (0.95, 0.95) \\
 & o4-mini & (0.49, 0.51) & 6.16\% & 6.64\% & 87.20\% & (0.49, 0.51) & (0.10, 0.12) & (0.92, 0.92) \\
 & Llama-3.1-8B (Distilled) & (0.81, 0.82) & 7.45\% & 6.39\% & 86.16\% & (0.86, 0.87) & (0.17, 0.19) & (0.92, 0.92) \\
 
\bottomrule
\end{tabular}

\caption{Alignment between LLM featurizers and human featurization obtained by majority voting for different models and assessment areas. QWK and F1 values presented are 95\% Bootstrap CI.}

\label{tab:alignment}

\end{table*}

}
\begin{table}[h]
    \small
    \centering
    \begin{tabular}{lc}
        \toprule
        \textbf{Assessment Area} & \textbf{Krippendorff's $\alpha$}\\
        \midrule
        Science & $(0.718, 0.723)$ \\
        Reading (Informational Text) & $(0.696, 0.700)$ \\
        Reading (Literature) & $(0.672, 0.680)$ \\
        \bottomrule
    \end{tabular}

    \caption{Inter-rater reliability among annotators for the featurization alignment experiment (95\% bootstrap CI).}
    % \caption{Krippendorff's $\alpha$ among human raters for the featurization alignment experiment (95\% bootstrap CI).}
    \label{tab:alpha}
\end{table}
\subsection{Scoring Accuracy Results}

% --- one-off starred footnote ---
\begingroup              % keep the change local
  \renewcommand{\thefootnote}{\fnsymbol{footnote}} % switch to symbol set (*, †, ‡ …)
% Table~\ref{tab:results} displays the results of the scoring accuracy experiments. Across items and within each assessment area, \textsc{AnalyticScore} outperforms several automated scoring baselines on average and, given its interpretability, achieves reasonable performance compared to state-of-the-art black-box models. Excepting the untuned Llama featurizer, each \textsc{AnalyticScore} variant outperforms\footnotemark[1] the few-shot prompting and automated scoring baselines. This highlights the striking improvement\footnotemark[1] from the raw Llama-3.1-8b-Instruct to the distilled model, with an average increase of 0.13 QWK. Distillation led to an especially notable increase in QWK for Science items (+0.19) and to performance comparable to both \textsc{AnalyticScore} GPT-4.1-mini featurizers. Compared to the best-performing models in each assessment area, these three \textsc{AnalyticScore} models are, on average, within 0.04 QWK for Science, 0.08 QWK for Reading (Informational Text), and 0.10 QWK for Reading (Literature) items.

% The data as a whole highlight the need to prioritize explainability. No model, including uninterpretable SOTA models, reliably met human baselines in performance. However, \textsc{AnalyticScore} models displayed remarkable results--0.06 QWK from the best performing fine-tuned LLM baseline-- while maintaining principled interpretability. Leveraging interpretable \textsc{AnalyticScore} models, we provide an operationalizable path forward as researchers advance LLM capabilities to encompass autoscoring needs.

Table~\ref{tab:results} shows the results of the scoring accuracy experiments. 
Across items and within each assessment area, \textsc{AnalyticScore} outperforms\footnotemark[1] several automated scoring baselines on average and, given its interpretability, achieves reasonable performance compared to state-of-the-art black-box models. Except for the untuned Llama featurizer, each \textsc{AnalyticScore} variant outperforms\footnotemark[1] the few-shot prompting and automated scoring baselines. Compared to the best-performing models in each assessment area, \textsc{AnalyticScore} is, on average, within 0.06 QWK for all items, 0.04 QWK for Science, 0.08 QWK for Reading (Informational Text), and 0.09 QWK for Reading (Literature) items. These results are noteworthy given that the FGTI principles strictly limit architectural flexibilities that often improve performance, such as subroutines that cannot be mapped to discrete reasoning steps or scoring components that are only partially articulable.

% For area-specific performance, the average performance of each of the three aforementioned \textsc{AnalyticScore} models is higher than that of the automated scoring baselines, and is higher or comparable to few-shot prompting across all assessment areas. 

Also noticeable is the striking improvement\footnotemark[1] in the performance of the Llama featurizer post-distillation, with an average increase of 0.13 QWK. The distilled Llama featurizer performs comparably to both variants of GPT-4.1 mini. Increase in average QWK is most notable for Science items (+0.19), followed by Reading (Literature) (+0.12) and Reading (Informational Text) (+0.05). 

% Overall, no SOTA models have achieved near-human performance. This highlights the need for model interpretability \todo

% Turning to different implementations of \textsc{AnalyticScore}, we observe that the GPT-4.1-mini featurizer with first-to-3 aggregation yields comparable performance with the deterministic variant. Yet, there is a striking improvement\footnotemark[1] from the raw Llama-3.1-8b-Instruct to the post-distillation model, with an average increase of 0.13 QWK. Increase in average QWK is most notable for Science items (+0.19), followed by Reading(Literature) (+0.12) and Reading(Informational Text) (+0.05). Overall, the distilled featurizer performs comparably to both GPT-4.1-mini featurizers, demonstrating the effectiveness of featurizer distillation.

\footnotetext[1]{$p<0.05$ for Wilcoxon signed-rank test across all items. Due to small $n$, no area-specific difference was statistically significant.}

\endgroup
% --- back to normal automatically ---

\subsection{Featurization Alignment Results}
\label{sec:results_alignment}

Table~\ref{tab:alpha} displays the Krippendorff's $\alpha$
% \footnote{$\alpha$ ranges between -1 and 1. 0 indicates chance agreement.} 
measured among the human raters in conducting the featurization task from Section~\ref{sec:featurize}. For all assessment areas, we observe $0.667\leq \alpha < 0.8$. These values fall within an acceptable range of inter-rater reliability \cite{krippendorff2018content}. We interpret this as a good level of rater agreement on the featurization process defined in this work
% , especially relative to similar rating tasks \cite{jurenka_towards_2024}, 
and acknowledge that there is still potential to refine and improve the task further.

Next, Table~\ref{tab:alignment} shows alignment between models and human labels. Most notably, the distilled Llama featurizer achieves substantially high agreement with human features across all assessment areas. Other featurizers also achieve high agreement in Science and Reading (Informational Text) but achieve moderate agreement in Reading (Literature). 

F1 scores and label distribution for each feature label\footnote{\label{distribution}For Human and o4-mini, we applied prevalence weighting to the 50 study samples to estimate the label distribution across all $(r,c)$ pairs.} provide a more detailed insight and reveal areas for further improvement. Notice that the F1 score is exceptionally high (near or above 0.9) for label 0, and moderate-to-high (0.6$\sim$0.93) for label 2, with higher agreement for Science items. Yet, alignment for label 1 is moderate-to-low, ranging from 0.68 down to 0.11. We attribute this result to the relatively ambiguous nature of the label category 1, coupled with the rarity of label 1 in human rating. While LLM featurizers achieve high overall alignment with human featurization, future work should reduce ambiguity in the featurization task and ensure model predictions better reflect the distribution of human featurization behavior.

\section{Conclusion}

Despite a pressing need, the AI and education research community has yet to develop a practical interpretability framework for automated scoring in large-scale educational assessments. In this work, we presented a principled approach to address this challenge. We analyzed the needs and potential benefits of assessment stakeholders and developed four foundational principles of interpretable automated scoring. As a baseline framework for future research, we developed the \textsc{AnalyticScore} framework. In the domain of text-based open-ended response scoring, \textsc{AnalyticScore} achieves promising scoring accuracy and demonstrates featurization behaviors that align with human judgment. We hope this work illuminates exciting new directions in developing practical and effective interpretable automated scoring methods for large-scale educational assessments.

\section*{Limitations} 

\paragraph{Capturing Complex Scoring Logic} While the average performance gap of 0.06 QWK between \textsc{AnalyticScore} and the uninterpretable SOTA is meaningful given the architectural constraints imposed by the FGTI principles, there is still room to improve. One potential area of improvement is the choice of the scoring module used during Phase 3 (Section~\ref{sec:scoring}). Our demonstration of \textsc{AnalyticScore} uses ordinal logistic regression as the traceable and interchangeable scoring module. While this implementation provides a strong baseline, it may not sufficiently capture the logical nuances required for scoring complex items. Future work should explore more complex yet traceable and interchangeable alternatives to ordinal logistic regression. Examples of possible alternatives are LLM workflows or agents which consist of modules that each compute a specific evidentiary reasoning step and human-understandable intermediate outputs.

\paragraph{Limited Benchmark Datasets} The ASAP-SAS dataset is the largest publicly available dataset of complex text-based constructed-response scoring, but it only reflects a tiny fraction of assessment items, response types, domains, test-taker population, and assessment language. For empirical results to robustly generalize, automated scoring methods should be comprehensively evaluated on various datasets collected across different assessment settings. Given the limited amount of publicly available large-scale benchmarks for complex open-ended response scoring, curation of additional datasets is necessary to support a more rigorous study of automated scoring.

\paragraph{Real-World Validation} Our analysis of assessment stakeholder needs and interpretability principles was scoped to a theoretical study derived from the educational assessment literature. To identify and address the gaps that exist between theory and practice, field studies should be conducted in the future to validate the needs of the real-world assessment stakeholders and iteratively refine the design of interpretable scoring systems.

\paragraph{Measuring Alignment} Our featurization alignment study (Sections~\ref{sec:eval_alignment} and \ref{sec:results_alignment}) measures the model-to-human and human-to-human alignment in featurization behaviors. In addition to measuring alignment in featurization behaviors, it is also important to measure how the extracted analytic components and the featurization task align with the target constructs that the assessment item is intended to capture \cite{bejar2016automated}. Analyzing construct alignment requires the knowledge of the design decisions involved in the development of the assessment items, which we did not have access to in our current study. We hope to see larger-scale studies on both featurization alignment and construct alignment in authentic assessment environments in the future.

\section*{Ethical Considerations}

\paragraph{Responsible Use of Model Interpretations}
To ensure that the assessment outcomes are valid, reliable, and fair, the use of automated scoring should be accompanied by the analysis of multiple sources of validity evidence \cite{bennett1998validity,bennett2015validity,williamson2012framework,bejar2016automated}. These sources include the features used by the scoring model, the model's alignment with human judgments, how the model handles unusual or unexpected responses, and the consistency of scoring behavior across student populations \cite{bennett2015validity}. 

Model interpretations are a means to an end for enabling these analyses, but having an interpretable scoring system does not, on its own, ensure that the assessment outcomes will be valid, reliable, and fair. Without a thoughtfully designed operational practice around the use of model interpretations, these systems have the risk of creating a \textit{false sense} of validity and trustworthiness when no proper auditing or mitigation is actually taking place. Beyond the technical and architectural aspects of interpretable automated scoring discussed in this work, it is also important to develop rigorous workflows, protocols, and systematic practices for using model interpretations to ensure that automated scoring systems are responsibly audited, monitored, and improved over time.

\paragraph{Student Privacy} Our demonstration of \textsc{AnalyticScore} involves the use of proprietary LLMs for extracting analytic components, featurizing responses, and creating the training dataset to distill the featurizer. When handling real-world assessment responses, care should be taken to avoid leakage of potentially sensitive or personally identifiable information contained in student responses.

\bibliography{references}

\appendix

\section{Prompts}
\label{sec:appendix_prompt}

\begin{figure}[h!]
    \centering
    \includegraphics[width=\linewidth]{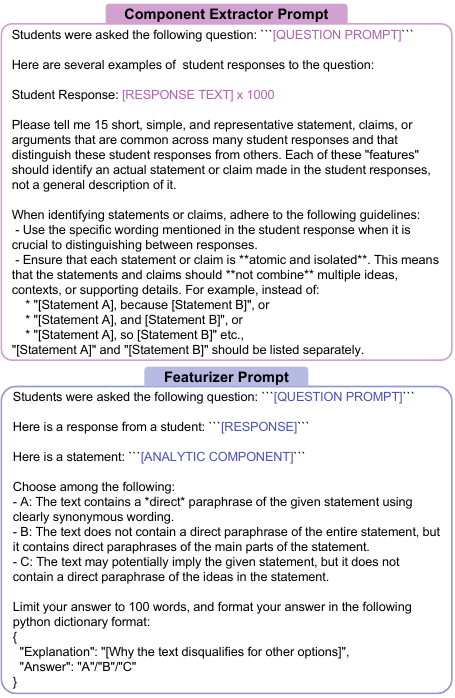}
    \caption{Prompts used in \textsc{AnalyticScore}}
    \label{fig:prompts}
\end{figure}

Figure~\ref{fig:prompts} shows the exact prompts used to implement the feature labeling function from Section~\ref{sec:featurize}.

\section{ASAP-SAS Dataset Detail}
\label{sec:appendix_asap}

Table~\ref{tab:dataset} shows the details of the ASAP-SAS dataset.

{
\setlength{\tabcolsep}{1mm}

\begin{table}[h!]
\centering
% \begin{tabular}{llllllllll}
% \toprule
% Q1 & Q2 & Q3 & Q4 & Q5 & Q6 & Q7 & Q8 & Q9 & Q10 \\
% \midrule
% 1341 & 1024 & 1445 & 1308 & 1459 & 1418 & 1432 & 1446 & 1453 & 1314 \\
% 331 & 254 & 363 & 349 & 336 & 379 & 367 & 353 & 345 & 326 \\
% 557 & 426 & 406 & 295 & 598 & 599 & 599 & 599 & 599 & 546 \\
% \bottomrule
% \end{tabular}

% \begin{table*}[h]
\small
\begin{tabular}{cccccc}
\toprule
& \textbf{Token Len.}
& \textbf{Train}
& \textbf{Valid}
& \textbf{Test}
& \textbf{Assessment Area}\\
\midrule
Q1 & $47.5\pm22.2$ & 1,341 & 331 & 557 & \multirow{2}{*}{Science}\\
Q2 & $59.2\pm22.6$ & 1,024 & 254 & 426 & \\
\midrule
Q3 & $47.9\pm14.6$ & 1,445 & 363 & 406 & \multirow{2}{*}{\makecell{Reading\\(Informational Text)}}\\
Q4 & $40.3\pm15.5$ & 1,308 & 349 & 295 & \\
\midrule
Q5 & $25.1\pm21.5$ & 1,459 & 336 & 598 & \multirow{2}{*}{Science}\\
Q6 & $23.8\pm22.6$ & 1,418 & 379 & 599 & \\
\midrule
Q7 & $41.3\pm25.1$ & 1,432 & 367 & 599 & \multirow{2}{*}{\makecell{Reading\\ (Literature)}}\\
Q8 & $53.0\pm32.6$ & 1,446 & 353 & 599 & \\
\midrule
Q9 & $49.7\pm36.3$ & 1,453 & 345 & 599 & \makecell{Reading\\ (Informational Text)}\\
\midrule
Q10 & $41.1\pm28.5$ & 1,314 & 326 & 546 & Science\\
\bottomrule
\end{tabular}

\caption{ASAP-SAS dataset detail by item.}
\label{tab:dataset}

% Q1 & 1341 & 331 & 557 & \multirow{2}{*}{\makecell{{Science} \\ (List of scientific procedure)}}\\
% Q2 & 1024 & 254 & 426 & \\
% \midrule
% Q3 & 1445 & 363 & 406 & \multirow{2}{*}{\makecell{Reading (Informational Text) \\ (Analysis + Evidence from text)}}\\
% Q4 & 1308 & 349 & 295 & \\
% \midrule
% Q5 & 1459 & 336 & 598 & \multirow{2}{*}{\makecell{Science \\(List of factual elements)}}\\
% Q6 & 1418 & 379 & 599 & \\
% \midrule
% Q7 & 1432 & 367 & 599 & \multirow{2}{*}{\makecell{Reading (Literature) \\ (Analysis + Evidence from text)}}\\
% Q8 & 1446 & 353 & 599 & \\
% \midrule
% Q9 & 1453 & 345 & 599 & \makecell{Reading (Informational Text) \\ (Analysis + Evidence from text)}\\
% \midrule
% Q10 & 1314 & 326 & 546 & \makecell{Science \\ (Reasoning + Data Anlysis)}\\
\end{table}
}

\section{Implementation Details}
\label{sec:appendix_implementation_details}

\paragraph{\textsc{ANALYTICSCORE} Implementation Details} We distilled the Llama featurizer for 2 epochs using a batch size of 8 and learning rate of 1e-4. All model calls were made through the official OpenAI API. Fine-tuning was conducted on an Ubuntu 20.04 machine with 2 RTX A6000 GPUs (49Gb memory), 16 AMD EPYC 9224 24-Core Processors, and 250Gb of CPU RAM.

\paragraph{Baseline Implementation Details}

Below is the list of hyperparameters used in this work. Lists indicate the range of hyperparameters searched.
\begin{itemize}
\item \textbf{Featurizer Distillation}

    \textbf{Seed:} 42;
	\textbf{Learning Rate}: 1e-4;
	\textbf{Epochs:} 2;
	\textbf{Batch Size:} 8;
	\textbf{Weight Decay:} 0.01;
	\textbf{Quantization:} 4-bit nf4;
	\textbf{LoRA Rank:} 16;
	\textbf{LoRA $\alpha$:} 8;
	\textbf{LoRA Dropout:} 0.05
\item \textbf{Few-Shot LLM}

	\textbf{Number of Fewshot examples:} 10 random few-shot examples per score category, sampled for each response;
	\textbf{Temperature:} 0.7
\item \textbf{SFT LLM (Encoder-Only Models)}

    \textbf{Seed:} 42;
    \textbf{Learning rate:} 3e-5;
    \textbf{Batch Size:} 12;
    \textbf{Early Stopping:} Threshold 1e-3, patience 10;
    \textbf{Classifier Head:} 2-layer feedforward network with hidden dim 32, dropout probability 0.1
\item \textbf{SFT LLM (Decoder-Only Models)}

    \textbf{Seed:} 42;
    \textbf{Learning Rate:} 1e-4;
    \textbf{Batch Size:} 8;
    \textbf{Epochs:} 20;
    \textbf{Early Stopping:} Threshold 1e-3, Patience 5;
    \textbf{Weight Decay:} 0.01;
    \textbf{Quantization:} 8-bit;
    \textbf{LoRA Rank:} 16;
    \textbf{LoRA $\alpha$:} 8;
    \textbf{LoRA Dropout:} 0.05
\item \textbf{AutoSAS}

    \textbf{Maximum Depth:} [50, 75, 100, 150, 200];
    \textbf{Number of Estimators:} [50, 75, 100, 150, 200];
    \textbf{Hyperparameter selection criteria:} Best QWK on validation set

    \textbf{Final selected hyperparameters:}
        \begin{description}
        \item[Q1.] \textbf{Max Depth}: 200; \textbf{\# of Estimators}: 150
        \item[Q2.] \textbf{Max Depth}: 200; \textbf{\# of Estimators}: 75
        \item[Q3.] \textbf{Max Depth}: 75; \textbf{\# of Estimators}: 100
        \item[Q4.] \textbf{Max Depth}: 75; \textbf{\# of Estimators}: 50
        \item[Q5.] \textbf{Max Depth}: 150; \textbf{\# of Estimators}: 100
        \item[Q6.] \textbf{Max Depth}: 50; \textbf{\# of Estimators}: 150
        \item[Q7.] \textbf{Max Depth}: 200; \textbf{\# of Estimators}: 150
        \item[Q8.] \textbf{Max Depth}: 200; \textbf{\# of Estimators}: 150
        \item[Q9.] \textbf{Max Depth}: 150; \textbf{\# of Estimators}: 200
        \item[Q10.] \textbf{Max Depth}: 200; \textbf{\# of Estimators}: 50
        \end{description}
\item \textbf{AsRRN}

    Parameters were mostly adopted from the following public implementation: \url{https://github.com/nkzhlee/AsRRN}

    \textbf{Seed:} 23;
    \textbf{Batch size:} 1;
    \textbf{Learning rate:} 1e-5;
    \textbf{Epochs:} 23;
    \textbf{LR gamma:} 0.1;
    \textbf{LR step:} 15;
    \textbf{Max sequence length:} 128;
    \textbf{Validation split:} 0.2;
    \textbf{Weight decay:} 0.0001;
    \textbf{Max gradient norm:} 1.0;
    \textbf{Warmup steps:} 0.2;
    \textbf{Gradient accumulation steps:} 1;
    \textbf{Pretraining steps:} 5;
    \textbf{Hidden dimension:} 768;
    \textbf{Message dimension:} 128;
    \textbf{Hidden dropout probability:} 0.2;
    \textbf{Number of graph steps:} 2;
    \textbf{Contrastive loss lambda:} 0.01;
    \textbf{Contrastive temperature:} 1;
    \textbf{Norm:} 1
\item \textbf{NAM}
    \textbf{N-gram sizes:} 2$\sim$10-grams
\end{itemize}

\section{Alignment Study Details}
\label{sec:appendix_alignment_study}

\begin{figure*}[h]
    \centering
    \includegraphics[width=0.8\linewidth]{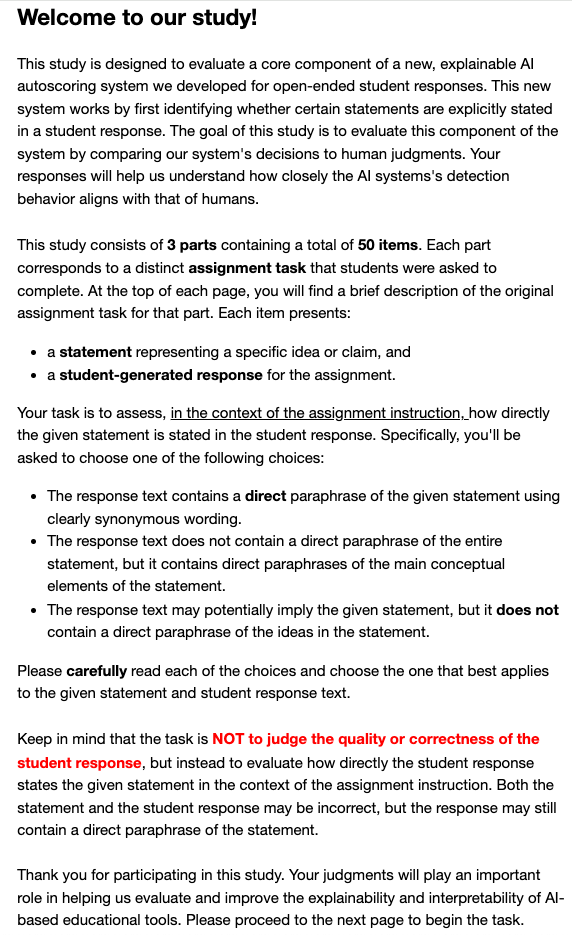}
    \caption{Welcome page of one of the 3 Qualtrics forms used for the featurization alignment experiment. This page contains explanations about the purpose of the study, the necessary contexts, and task description.}
    \label{fig:survey_purpose}
\end{figure*}

\begin{figure*}[h]
    \centering
    \includegraphics[width=0.8\linewidth]{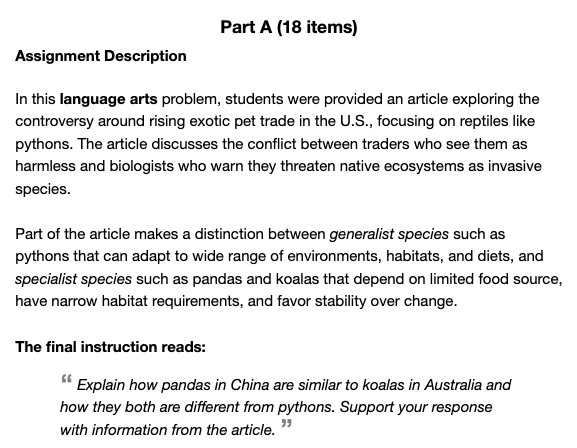}
    \caption{Example description of the assessment item and assessment instruction shown to the annotators.}
    \label{fig:survey_asgn_task}
\end{figure*}

\begin{figure*}[h]
    \centering
    \includegraphics[width=0.8\linewidth]{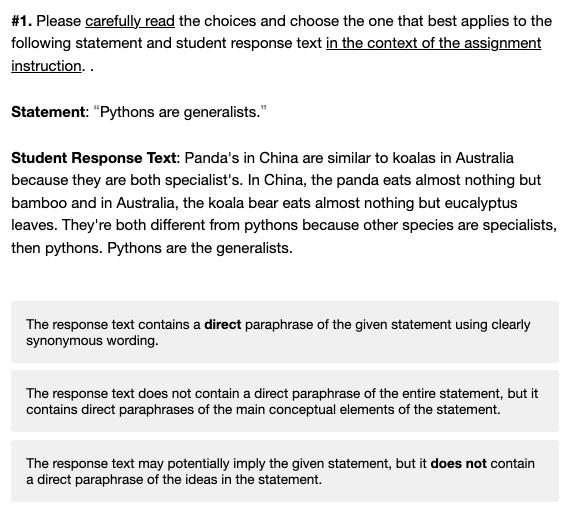}
    \caption{Each annotation task presented annotators with a (response, analytic component) pair and asked them to select one of the three label options used in \textsc{AnalyticScore}'s labeling function.}
    \label{fig:survey_feat_task}
\end{figure*}

The annotators received an oral presentation of the purpose of the study along with links to 3 Qualtrics forms to be filled out, one in each assessment area. The form reiterated the study’s purpose, explained the task, and presented 50 items to annotate, each containing the context of the assessment item and the same featurizer prompt shown in Figure~\ref{fig:prompts}. The overall process took each annotator between 2.5 and 3.5 hours. Figures~\ref{fig:survey_purpose} through \ref{fig:survey_feat_task} are example screenshots from one of the three Qualtrics forms used for annotation.

\end{document}